\pdfoutput=1
\documentclass[conference]{IEEEtran}
\IEEEoverridecommandlockouts
\usepackage{cite}
\usepackage{graphics} 
\usepackage{graphicx}

\usepackage[bookmarks=false]{hyperref}
\hypersetup{
    colorlinks=true,
    linkcolor=black,
    filecolor=black,
    citecolor=black,
    urlcolor=blue,
}
\usepackage[utf8]{inputenc}
\usepackage{listings}
\usepackage{subcaption}
\usepackage{algorithmic}
\usepackage[]{units}
\usepackage{pgfplots}
\pgfplotsset{compat=1.14}
\usepackage{xcolor}
\usepackage{siunitx}
\usepackage{multirow}
\graphicspath{ {Pictures/} }
\usepackage{amsmath,amssymb,amsfonts}

\usepackage[acronym,nomain,nonumberlist]{glossaries}
\newacronym{mav}{MAV}{Micro Aerial Vehicle}
\newacronym{mlp}{MLP}{MultiLayer Perceptron}
\newacronym{ar}{AR}{Augmented Reality}
\newacronym{ros}{ROS}{Robot Operating System}
\newacronym{gps}{GPS}{Global Positioning System}
\newacronym{vi}{VI}{Visual Inertia}
\newacronym{fps}{fps}{Frame Per Second}
\newacronym{mae}{MAE}{mean absolute error}
\newacronym{dof}{DoF}{Degree of Freedom}
\newacronym{ekf}{EKF}{Extended Kalman Filter}
\newacronym{ahe}{AHE}{Adaptive Histogram Equalization}
\newacronym{pid}{PID}{Proportional Integral Derivative}
\newacronym{mimo}{MIMO}{Multiple Input Multiple Output}
\newacronym{rl}{RL}{Reinforcement Learning}
\newacronym{vo}{VO}{Visual Odometry}
\newacronym{slam}{SLAM}{Simultaneous Localization And Mapping}

\newlength\fwidth

\begin{document}

\title{A Subterranean Virtual Cave World for Gazebo based on the DARPA SubT Challenge
\thanks{This work has been funded by the European Unions Horizon 2020 Research and Innovation Programme under the Grant Agreement No. 730302 SIMS. \newline Corresponding author's email: antkov@ltu.se
}
}
\author{\IEEEauthorblockN{Anton Koval, Christoforos Kanellakis, Emil Vidmark, Jakub Haluska and George Nikolakopoulos}
\IEEEauthorblockA{
Robotics Team\\ Department of Computer Science, Electrical and Space Engineering\\ Lule\r{a} University of Technology \\ Lule\r{a} SE-97187, Sweden
}
}

%

%

%
\maketitle              
\begin{abstract}
Subterranean environments with lots of obstacles, including narrow passages, large voids, rock falls and absence of illumination were always challenging for control, navigation, and perception of mobile robots. The limited availability and access to such environments restricts the development pace of capabilities for robotic platforms to autonomously accomplish tasks in such challenging areas. The Subterranean Challenge is a competition focusing on bringing robotic exploration a step closer to real life applications for man-made underground tunnels, urban areas and natural cave networks, envisioning advanced assistance tools for first responders and disaster relief agencies. The challenge offers a software-based virtual part to showcase technologies in autonomy perception, networking and mobility for such areas. Thus, the presented open-source virtual world aims to become a test-bed for evaluating the developed algorithms and software and to foster mobile robotics developments. 
\end{abstract}
\glsresetall
\section{Introduction}
Defence Advanced Research Projects Agency (DARPA)~\cite{darpa} was always pushing robotic technologies to high technology readiness levels. Back in 2004, DARPA run path breaking Grand Challenge~\cite{grandchallenge} with the goal to accelerate the development of autonomous vehicle technologies. And now, we are seeing, that solved challenges by the research community, allowed to create commercial self-driving cars and other ground vehicles.

Nowadays, robotic technologies are constantly evolving and getting integrated in industrial applications. Presently, they closely interact with human operators in space, mining, search and rescue and other hazardous applications. This makes research towards mobile robots' development and advancement of their autonomy levels of high interest~\cite{siciliano2016springer}. However, still exist some factors that limit these developments. For instance, one of them is restricted access to underground mine tunnels, voids or shafts due to production process, another one is that risk of losing an expensive robot with no possibility to recover it. 
Current technologies fail to provide persistent situational awareness of the diverse subterranean operating environments, including tunnels, urban underground, and cave networks. Therefore, the DARPA Subterranean (SubT) challenge~\cite{subtworld} motivates the development of novel approaches and technologies to allow warfighters and first responders to rapidly map, navigate, and search dynamic underground environments.

\begin{figure}[htbp!]
  \centering
  \includegraphics[width=\linewidth]{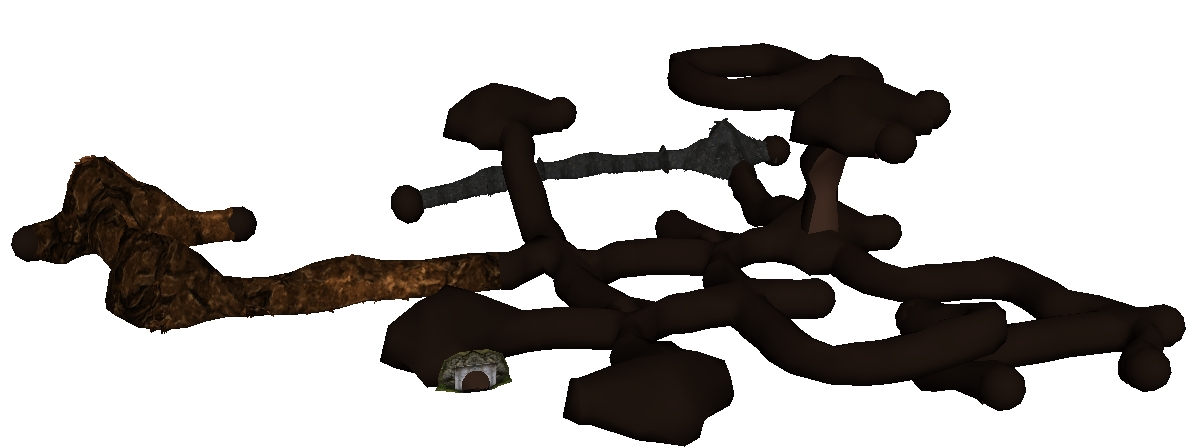} 
  \caption{Perspective view of the cave world, while its video demonstration is available at~\url{https://youtu.be/h06Xb5bNiGs}}
  \label{fig:cave_world}        
\end{figure}

Thus, to accelerate the development pace of robotics, simulation approaches are of significant importance. Nowadays, the most widely used framework for robotics developments is \gls{ros}~\cite{quigley2009ros} which provides free and open source tools, peer-to-peer communication and combines developments for most of existing robots, sensors and actuators. Moreover \gls{ros} provides all necessary interfaces to simulate robot perception and dynamics in Gazebo 3D Simulator~\cite{gazebo}, which is commonly used for evaluation of developed algorithms and software. To this end, we introduce an open-source virtual caving world, as depicted in Figure~\ref{fig:cave_world}, for Gazebo and~\gls{ros} that is available at~\url{https://github.com/LTU-CEG/gazebo_cave_world} and is described in the following Section~\ref{world_description}.
\section{Description}
\begin{figure}[htbp]
  \centering
  \begin{subfigure}[b]{0.35\linewidth}
    \centering
    \includegraphics[scale=0.3]{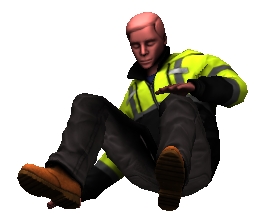}
    \caption{ }
    \label{fig:survivor}
  \end{subfigure}
  \qquad
  \begin{subfigure}[b]{0.2\linewidth}
    \centering
    \includegraphics[scale=0.3]{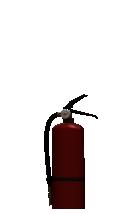}
    \caption{ }
    \label{fig:extinguisher}
  \end{subfigure}
  \qquad
  \begin{subfigure}[b]{0.25\linewidth}
    \centering
    \includegraphics[scale=0.3]{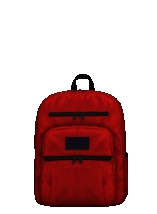}
    \caption{ }
    \label{fig:backpack}
  \end{subfigure}
  \caption{Examples of artifacts: (\subref{fig:survivor}) - survivor, (\subref{fig:extinguisher}) - fire extinguisher and (\subref{fig:backpack}) - backpack}
  \label{fig:artifacts}
\end{figure}

{\label{world_description}}
\begin{figure*}
  \centering
  \begin{subfigure}[b]{0.355\textwidth}
    \includegraphics[width=\textwidth]{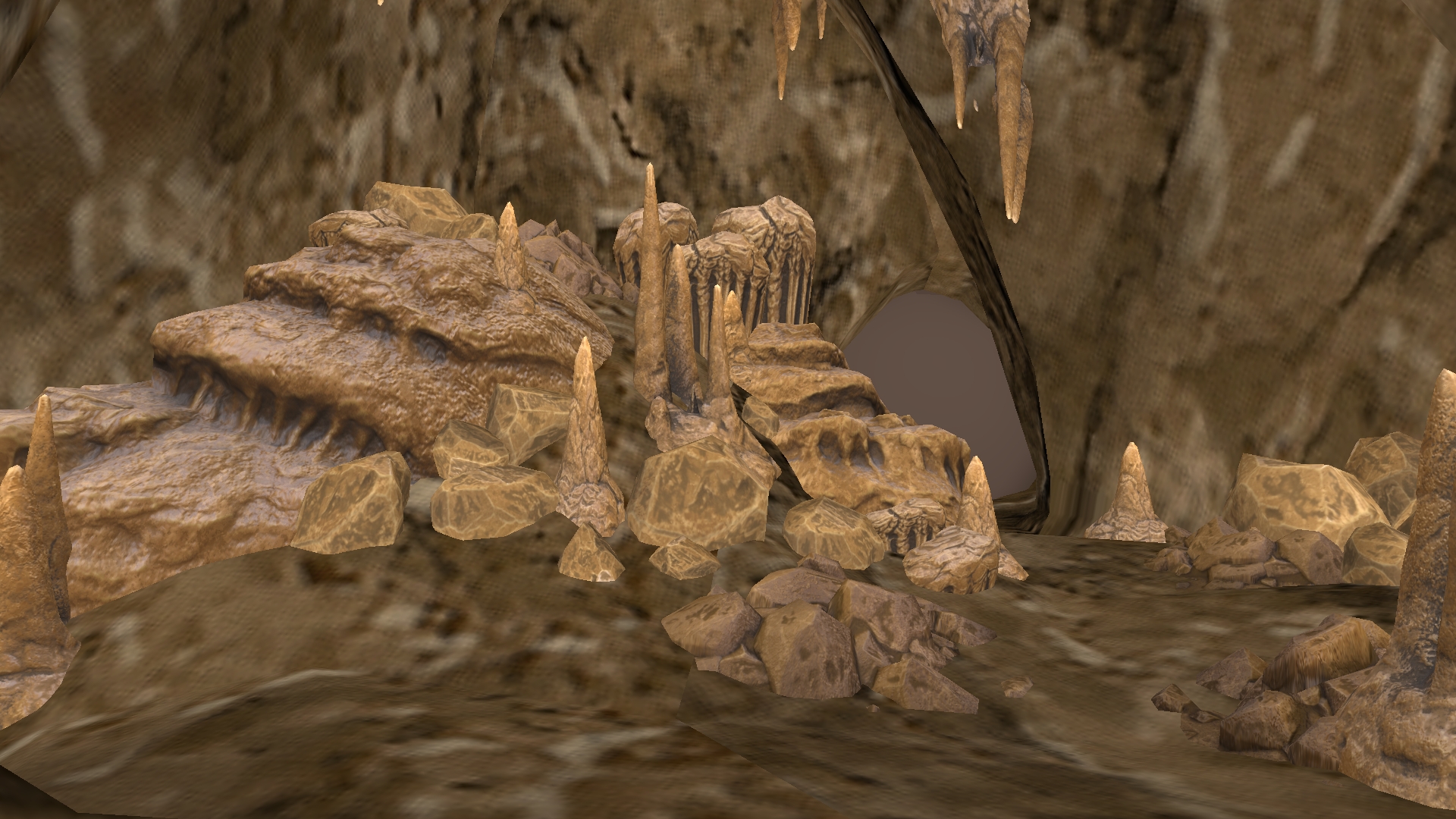}
    \caption{ }
    \label{fig:void}
  \end{subfigure}
  \qquad
  \begin{subfigure}[b]{0.355\textwidth}
    \includegraphics[width=\textwidth]{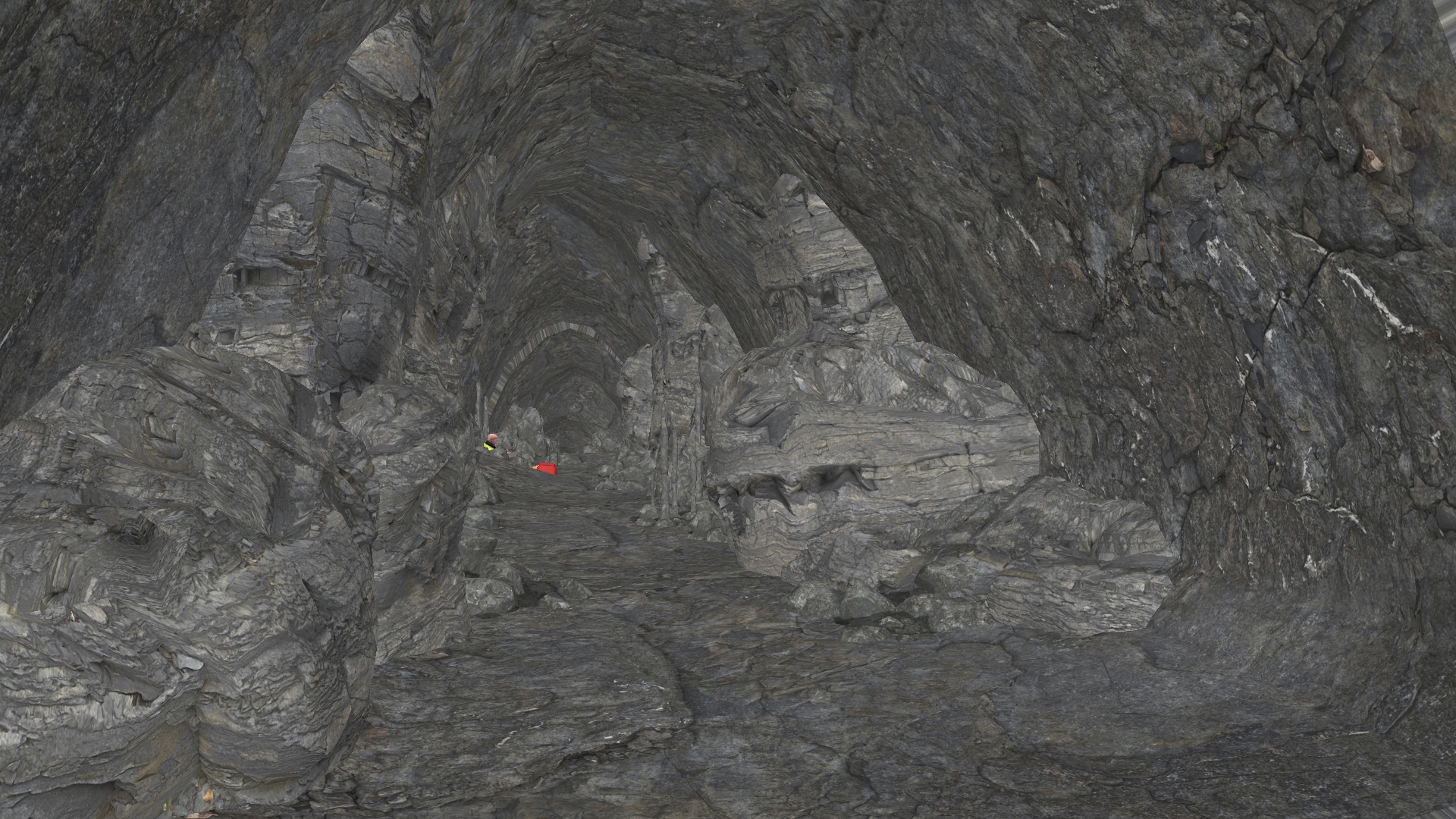}
    \caption{ }
    \label{fig:tunnel}
  \end{subfigure}
  \qquad
  \begin{subfigure}[b]{0.355\textwidth}
    \includegraphics[width=\textwidth]{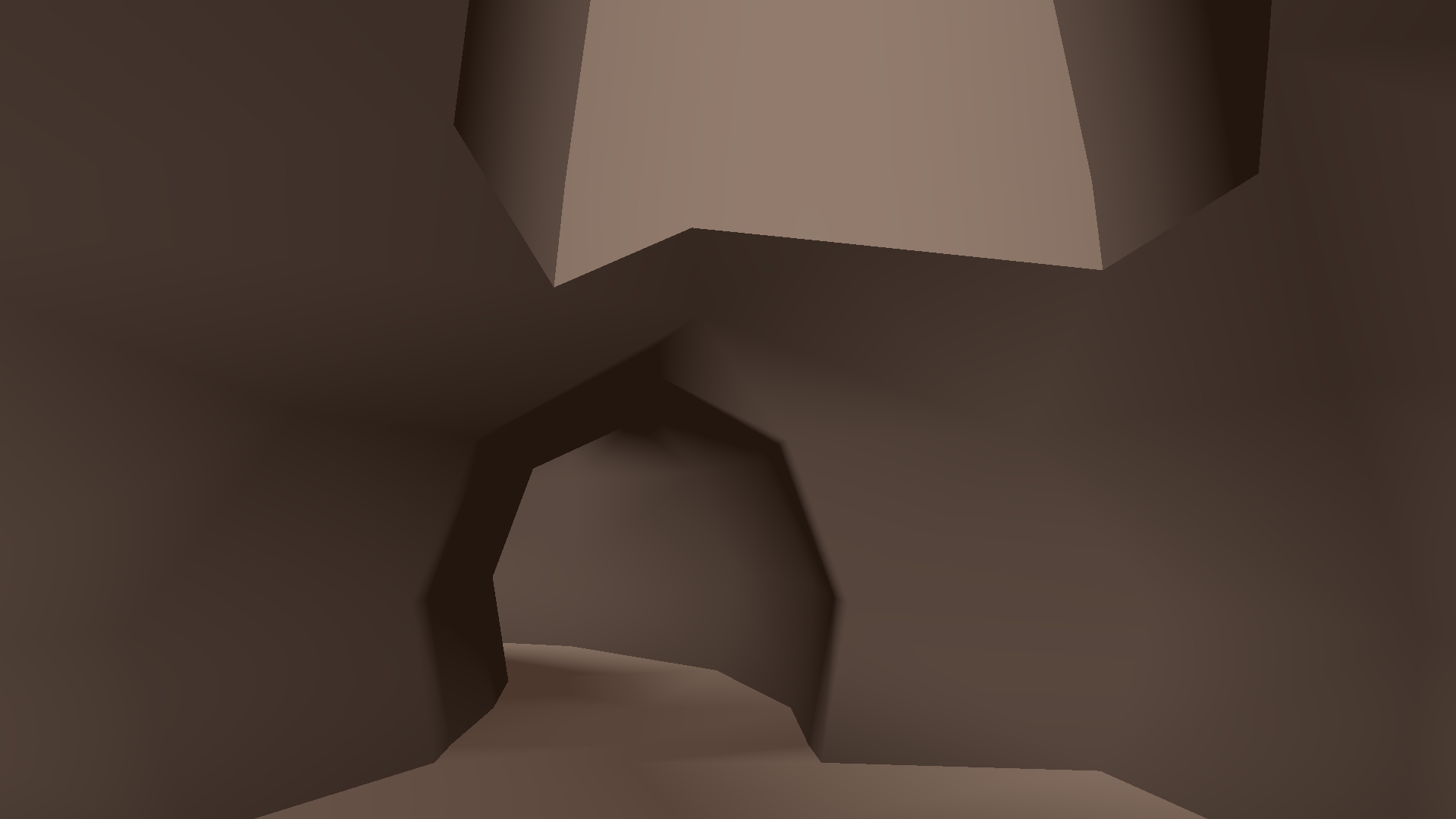}
    \caption{ }
    \label{fig:void2}
  \end{subfigure}
  \qquad
  \begin{subfigure}[b]{0.355\textwidth}
    \includegraphics[width=\textwidth]{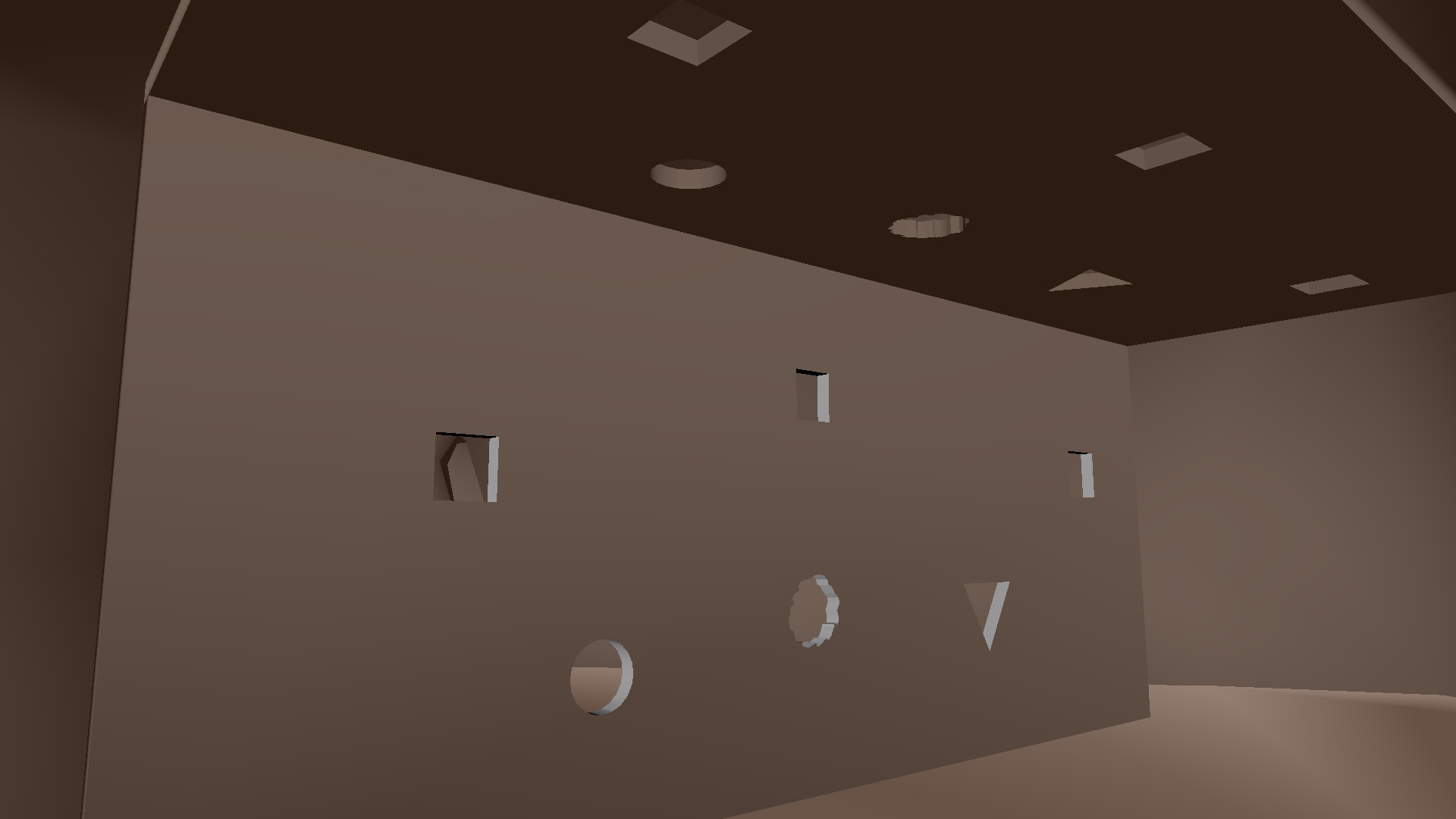}
    \caption{ }
    \label{fig:openings}
  \end{subfigure}
  \caption{Examples of terrains in the virtual world: (\subref{fig:void}) - void area with stalactites and stalagmites, (\subref{fig:tunnel}) - tunnel with rockfalls,  (\subref{fig:void2}) - void with vertical shaft and (\subref{fig:openings}) - foldable drone evaluation area}
  \label{fig:terrains}
\end{figure*}
%
%
The virtual cave environment (Figure~\ref{fig:cave_world}), is made by using both DARPA provided tiles~\cite{subtworld} and custom made 3D models with the following features:
\begin{itemize}
    \item Large void areas, as depicted in Figure~\ref{fig:terrains}\subref{fig:void};
    \item Tunnels, ramps going up and down (Figure~\ref{fig:terrains}\subref{fig:tunnel}) and vertical shafts (Figure~\ref{fig:terrains}\subref{fig:void2}) with average opening size of 5.3 x 4.5 [meters];
    \item Rockfalls;
    \item Gate with apriltags~\cite{wang2016apriltag} that provide ground truth information;
    \item Foldable unmanned aerial vehicle evaluation area (Figure~\ref{fig:terrains}\subref{fig:openings}), which has openings of various size and shape;
    \item Areas with stalactites and stalagmites;
    \item Possibility of placement of artifacts - objects of interest that could be found and localised. Their examples are depicted in Figure~\ref{fig:artifacts};
    \item Various textures;
    \item External dimensions of the world are (height x width x length): 33 x 102 x 150 [meters]
\end{itemize}
%
%
The designed virtual world offers realistic caving environment for rapid algorithm testing and evaluation for all kinds of ground and aerial robots and their groups with different levels of autonomy. It is designed to:
\begin{itemize}
    \item Test localization, exploration, obstacle avoidance and path planning algorithms.
    \item Test object detection and localization algorithms.
    \item Train AI systems.
    \item Perform regression testing.
\end{itemize}
\section{Developments}{\label{developments}}
The virtual cave world is a seamless model in $obj$ format. Its caving system contains 17 branches, while the topology of the world includes ramps going up and down, vertical shafts and blockers. The world is designed in the way that blocker elements can be removed in order to extend it with areas of interest.
\section{Citing This Work}{\label{citing_this_work}}
This open source virtual world is published under the MIT License. However, if you use the world we appreciate that you cite it accordingly.
For academic publications, you can cite as follows:
%
\begin{lstlisting}
  @misc{akoval2020,
  author = {Anton Koval and Christoforos Kanellakis
  and Emil Vidmark and Jakub Haluska and
  George Nikolakopoulos},
  title = {A Subterranean Virtual Cave World
  for Gazebobased on the DARPA SubT Challenge},
  year = {2020},
  publisher = {arXiv},
  howpublished = 
  {\url{http://arxiv.org/abs/2004.08452}}}
\end{lstlisting}


%
\bibliographystyle{IEEEtran}
\bibliography{mybib}
\end{document}